\documentclass{article}
\usepackage{arxiv}
\usepackage{graphicx}
\usepackage[utf8]{inputenc} 
\usepackage[T1]{fontenc}    
\usepackage{hyperref}       
\usepackage{url}            
\usepackage{booktabs}       
\usepackage{amsfonts}       
\usepackage{nicefrac}       
\usepackage{microtype}      

\title{LookALike: Human Mimicry based collaborative decision making}

\author{\href{https://orcid.org/0000-0002-6705-6506}{\includegraphics[scale=0.06]{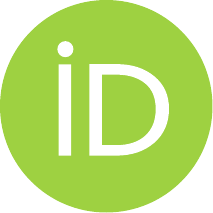}\hspace{1mm}Rabimba Karanjai}\thanks{www.rabimba.me} \\
	Department of Computer Science\\
	University Of Houston\\
	\texttt{rkaranjai@uh.edu} \\
        \And
	Weidong Shi \\
 	Department of Computer Science\\
        University Of Houston \\
	\texttt{wshi3@uh.edu}\\
}

\begin{document}
\maketitle
\begin{abstract}
  Artificial General Intelligence falls short when communicating role specific nuances to other systems. This is more pronounced when building autonomous LLM agents capable and designed to communicate with each other for real world problem solving. Humans can communicate context and domain specific nuances along with knowledge, and that has led to refinement of skills. In this work we propose and evaluate a novel method that leads to knowledge distillation among LLM agents leading to realtime human role play preserving unique contexts without relying on any stored data or pretraining. We also evaluate how our system performs better in simulated real world tasks compared to state of the art.
\end{abstract}




\maketitle

\section{Introduction}
The rise of large, pre-trained models has changed the AI landscape. Transformer-based language models (LLMs) excel at natural language tasks. Researchers are now trying LLMs for sequential decision-making – a core skill for AI agents. While LLMs can suggest basic plans \cite{huang2022language}, they often fail to consider real-world constraints or long-term planning.  To improve this, approaches using environmental feedback are being explored. This feedback could be sensory data \cite{xiang2024language}, human input \cite{yao2022react}, or information about the plan's progress \cite{raman2022planning}, allowing the AI to adjust its plans accordingly.

LLMs show promise as planners, but they're far from perfect.  A key hurdle is their reasoning ability; even given detailed instructions, they struggle to reliably produce functional plans \cite{silver2022pddl}. Current methods also force LLMs to learn through real-world (or simulated) trial and error, which is slow and costly. This differs from classical planning approaches \cite{fikes1971strips} where hypothetical plans can be analyzed for potential errors. Further, LLMs can be frustratingly stubborn, repeating the same mistakes despite feedback. Researchers are pushing the limits of what LLMs can learn and generalize, exploring new methods inspired by cognitive science to understand how information is shared within these models.

In this work, we explore different challenging reasoning scenarios involving both text based games and visual scenarios, more complex than previous studies. We explore a mimicking approach from the domain of cognitive science \cite{whiten1998imitation, dawson1965observational} and see if it can be combined with more traditional imitation learning approaches \cite{10.1609/aaai.v33i01.33019900}. And through that we try to answer if we can introduce context aware problem solving skills to LLMs that are not capable of it without the framework.

\section{Problem Statement}
The ability to gain knowledge, skills and solve relevant tasks and problems is generally defined as Intelligence\cite{chollet2019measure}. Such knowledge helps execute step by step tasks leading to a specific outcome. A lot of times these depend on a context of the task, to solve them effectively. Huamns gain these context and knowledge often from environment and other huamns, proving the context and knowledge transmissions as an effective tool in the context of problem solving. However in case of LLM's and LLM agents that is not the case always. The world knowledge, as available to us, is not available to a LLM making it much inefficient (and at times impossible) to solve certain problems. Knoweledge trasmission is a form of social learning, often assisted by other humans. This is common in everyday social interaction for humans. Corsss knoweldge transfers like copying a recipe by seeing someone cook, to learning how to play a game and learning rules of a game, are all cross domain knowledge transfer we do naturally. Which the LLMs fail to capture.

In our work, we seek to introduce our framework to capture the visual modal of these novel contexts to augment an LLMs understanding of a problem, and also cross transfer that to other agents preserving world knowledge for that task. 

\subsection{Motivation}

The motivations for this work are the following:
\begin{itemize}
    \item Increase domain specific reasoning capabilities of the LLM agents without data dependent training
    \item Rich visual scenes encode temporal and contextual knowledge that is beneficial and often crucial for solving a lot of real world tasks, which often is not encoded or captured by LLMs unelss specifically encoded.
    \item Contextual knowledge and knowledge transfer is a natural part of human intelligence. As such it is assumed they would also benefit AGI or to achieve AGI.
\end{itemize}

\section{Reasoning Playground Design}

\subsection{Tasks}
We specifically chose two tasks for our reasoning evaluation. One involving playing textual games based on world knowledge. Playing games has been one of the most evaluated tasks among the world planning scenarios \cite{tan2023text}. Artificial intelligence (AI) has made historic strides through games.  A major milestone was IBM's Deep Blue defeating chess champion Garry Kasparov in 1997\cite{campbell2002deep}. Google DeepMind's AlphaGo \cite{silver2016mastering} made history in 2016 by beating a professional Go player, a feat previously considered difficult for AI due to the game's complexity.  AI continued to conquer poker with DeepStack\cite{moravvcik2017deepstack} and Libratus\cite{brown2018superhuman},  programs that mastered heads-up no-limit Texas Hold'em in 2017. In 2019,  OpenAI Five and DeepMind AlphaStar made waves by defeating world-class players in the complex strategy games Dota 2 \cite{berner2019dota} and StarCraft II\cite{arulkumaran2019alphastar}, respectively. These victories underscore the increasing ability of AI to tackle increasingly challenging game environments.

To address the identified limitations~\cite{sobieszek2022playing} of the existing work . For our evaluation.  we chose to use Zork \cite{hausknecht2020interactive,tsai2023can}.

For a more comprehensive world planning we chose to utilize procgen \cite{cobbe2020leveraging} to generate world scenarios for task and see if our farmework achieves goal in defined time.

\section{Architecture}
\subsection{Learning}
We believe the ability to faithfully learn from others is a powerful tool for LLMs – and we show that reinforcement learning can develop this skill solely from seeking rewards using a reward model actor. Our work requires just a few basic assumptions: that the environment includes a domain expert to mimic, and that the AI agent itself has some minimal ability to analyze and store information. The challenge is that the AI must figure out how the skilled individual acts without direct access to their decision-making process. We utilize a reward model\cite{silver2021reward} mechanism to reward the AI agent to guide its actions towards favorable outcomes from the domain expert model. If our AI successfully learns to imitate, then the reinforcement learning process itself must have pushed for accurate, adaptable imitation that works in new situations.  We demonstrate this using maximum a posteriori policy optimisation\cite{abdolmaleki2018maximum}.

In reinforcement learning (RL), there's a constant dillemma between trying new things (exploration) and using what you already know (exploitation). Exploration is vital for finding better strategies, but it can be risky, especially when rewards are rare or the problem is really complex\cite{kearns2002near}. The risk is getting stuck doing something average instead of finding what works best.  We demonstrate how an AI can learn to model its environment, including the actions of others, to tackle these difficult exploration challenges.

The trick is that standard approaches to RL don't work when you can't see everything that's going on or when the system is too complex to perfectly map out. In these cases, the AI needs to learn not only for itself but also by cleverly figuring out what the 'domain experts' around it are doing.  We show that with the right setup, this kind of observational learning can happen through standard reinforcement learning, without any special instructions to imitate. This opens up new possibilities for solving truly challenging problems.

\subsection{Playground}

For Zork, we followed \cite{tsai2023can} closely to replicate the environment for playing the game. We took a similar approach to piping the question and answers from the game back to forth from the LLM. HOwever unlike their work, we have automated to process to keep make the experiment reproducible and run multiple occurrences.

For procgen we utilized their API to generate and solve the game models passing the task and task instructions through the LLM. 

\section{Evaluation Architecture}

Our system comprises of tow sets of LLMs. One is the \textit{player} LLM. (pLLM) which gets the environment information and comes up with the instructions to pass on to the game. This is true for both text based as well as vision based tasks. The player LLM (pLLM) in our case is gemma-7b\cite{gemma}. We intentionally chose a moderate sized LLM to see if our framework can actually utilize the encoded context information to get better apart from its own learned representations of knowledge.

The system also utilizes a separate Google Gemini 1.5 Pro model as a Domain Expert(vLLM). This model gets input from existing plays by human players to construct a reward mechanism for the pLLM. As an example, for zork, this gets input in few shot prompting of perfect Zork plays and extracts the gameplay information from the video as we see from Figure . Since Gemini 1.5 is capable of handling a very large context and also multimodal, this was perfectly suited for our automated domain expert critique. 

\begin{figure}
    \centering
    \includegraphics[width=1\linewidth]{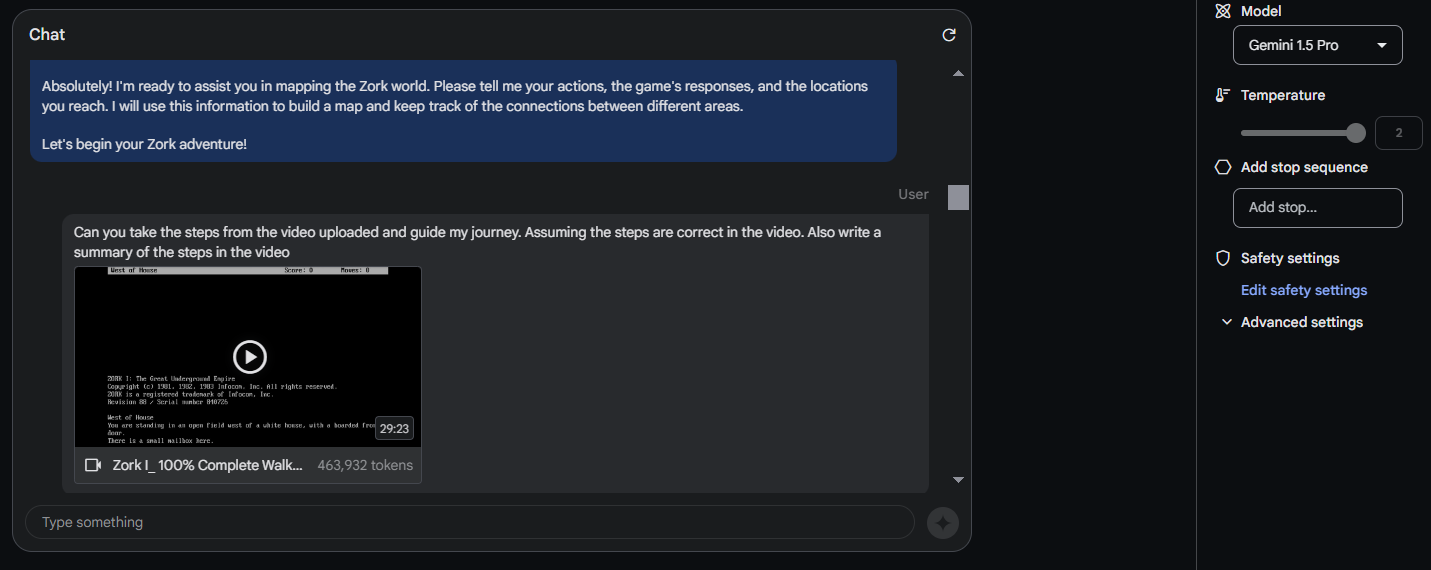}
    \caption{Zork Human Gameplay ingestion}
    \label{fig:gemini-zork}
\end{figure}

Once it creates a outline of the steps, it can generate an optimal pathway for the game to play and can reward/punish the pLLM based on the existing reward. This vLLM (Vision LLM) is also used for vision related data ingestion for procgen tasks.

\subsection{Constructing Game Play Solution}
If we generalize our task, our solution starts by feeding pLLM a variety of information. We start with detailed instructions on how to play the game. Example videos from similar tasks are fed to vLLM for building a reward model. Next, we provide context about the current situation, including the agent's goals and any game limitations. This is followed by a description of the specific action the agent is taking. Finally, we include a list of reusable terms so the LLM maintains a consistent language within the plan. Each of the steps is rewarded/punished based on the outcome as determined by the vLLM model.

An AI agent must first acknowledge the existence of other players \cite{ndousse2020multi} and the likelihood that they share its objectives and capabilities in order to use information from them successfully.  The agent won't, however, have direct access to the thoughts or acts of the other players. In order to solve this, we give the AI a unique "attention" mechanism that aids in keeping it focused on the locations of other players in its surroundings.  Similar to a spotlight, this attention method aids the AI in recognizing what's crucial, in this case, maintaining track of the other participants.  When the AI has mastered paying attention to the appropriate things, we no longer need the unique sensor that we used during training to assist it learn this skill. This is only done for pLLM, with different temperature for it to assume different role. Once pLLMs with different temperature goes through the rewards of vLLM, we take into account the most successful one. 

In case of pLLM swarms, for each step, after each reward/punishment for each of the pLLMs for each step, the results are stored and passed back as "memory" to the pLLMs along with the rewards/ punishment information for the next step, before being rewarded.

\section{Results}
We tested our approach in three settings: just asking a pLLM on the game play, and then on two settings, with vLLM reward and one with both vLLM reward as well as swarm input. POur swarm evaluation setup is more complex than other benchmarks, with the pLLM having more inputs to consider as piori than just a reward being given \textit{after} the decision.

Our experiments focused on a few key areas. First, we looked at how well the LLMs generated gameplay is.  Next, we tested if the LLMs could use collaborative feedback from piror steps to fix errors in their play. Since the pLLMs were run in different temperature (and intentionally not trying to be reproducible in their respective plays) the feedback from the reward models varied. We looked at if these were beneficial for another pLLM to imporve their game play assuming a different role. Finally, we demonstrated how the just giving vLLM produced rewards along with pLLM swarm could be used by the AI for actual planning and problem-solving. We used two language models for this: gemma-7b and Gemini 1.5 Pro.

\subsection{Plan Generation}
We started by testing how well the language models could come up with game plan when only some of the rules were provided. This is important because in real-world situations, you often know some constraints about the possible actions beforehand.  We focused on whether the models could accurately recreate a "correct" game plan that took those constraints and the relationships between actions into account. Our results show that without reward Gemini significantly outperforms gemma-7b in generating accurate plans. Table 1 details the number of errors for each domain, and to give you an idea of the complexity involved, we've also included the number of parameters and literals in the final, corrected Gemini plans. We can see the errors in Table \ref{tab:errors}.

\begin{table}[]
\resizebox{\columnwidth}{!}{%
\begin{tabular}{|l|l|l|l|}
\hline
GamePlay        & \# of actions & gemma-7b errors & gemini errors \\ \hline
Map Navigation  & 27            & 248+            & 17            \\ \hline
Decisive Action & 15            & 88+             & 6             \\ \hline
\end{tabular}%
}
\caption{Table \ref{tab:errors}: The number of errors in the gameplay steps produced by the LLMs.
A "+" mark means the generated model contains too much irrelevant information, which obscures the true number of errors.}
\label{tab:errors}
\end{table}

\subsection{Rewarding using vLLM}
We decided to focus on the reward models Gemini produced with only videos as input. Our goal was to show that Gemini could be used as a tool to take video and fix errors in its plans mimicking the human gameplay. These mimicking behaviours would act as reward mechanism for pLLMs swarm decision making collaboration.

By providing feedback like "Possible actions:
Open mailbox
Go north
Go south
Go west
Recommended action:
Open the mailbox." Gemini could pinpoint and define decisive actions. These action rewards successfully corrected a lot of incorrect outputs from the pLLMs and that in turn helped others in the swarm to reach the solution faster.

For different tasks we can see in Table \ref{tab:success} our method remarkably improves plan success rate, and solves Zork comfortably. Even for procgen tasks, which the standalone pLLM couldn't solve at all, show an increased success.

\begin{table}[]
\resizebox{\columnwidth}{!}{%
\begin{tabular}{|l|l|l|}
\hline
Player Type       & Zork & procgen \\ \hline
pLLM              & 35   & 0       \\ \hline
pLLM+vLLM         & 37   & 15      \\ \hline
pLLM Swarm + vLLM & 100  & 56      \\ \hline
\end{tabular}%
}
\caption{Table : Success rates of different approaches in solving a gameplay}
\label{tab:success}
\end{table}

\section{Discussion}

There are two perspectives on open-endedness in LLM systems for problem solving.  We can train an LLM to achieve a specific degree of success over a wide range of tasks that get harder and harder, or we can concentrate on one complex activity and aim to create an LLM that keeps getting better at it. For our use case—coming up with solutions for game-driven tasks—we found the second technique to be more useful.

Our framework itself is surprisingly basic. This shows that the way to create flexible, "collaborative" LLMs that can correct themselves from cooperative interactions may be to use fewer, more potent components. In the future, it will be essential to develop scalable training techniques, enhance how AIs learn about the world, and build systems in which the training process itself is dynamic.

Our method of using social learning to teach the LLM is similar to "memory-based" super-learning. As an example in the event that a new book in a library or if the catalogue shifts, you would want a librarian robot to be able to rapidly adapt. Sending all of its data to a massive server may also raise privacy issues. Our LLM absorbs human knowledge by emulating human behavior and storing all of it in its own memory. This maintains privacy while making it flexible.

We have demonstrated that social learning research can be conducted in complex visual contexts as well. But scale alone isn't the key. A scaled-down version of our setup could also provide exciting findings. The scalability of the approach is a powerful incentive for further research in this field, regardless of one's access to massive computing resources.

\section{Conclusion}

In this work we have introduced a new way of using language models (LLMs) for planning and solving tasks. Instead of having them generate plans directly, we use LLMs to build a model of how the world works by participating in a collaborative swarm and rewarding correct actions based on na domain expert(vLLM). This approach leverages the strength of LLMs – understanding the world – while avoiding their weakness in complex planning. Our process starts with Gemini generating a detailed vision guided reward plan. We then use this to refine the pLLMs. Finally, use this collaborative decision to play and evaluate the games. Each of these pLLMs assume a different role for a gamer, and achieves efficiency by collaborating and passing contextual information.

There are rooms for improvement in our system.  First, our test environments were still relatively simple compared to some planning challenges. Can LLMs scale to more complex logic? Second, we currently assume the the videos are enough for domain expert and procgen correctly generates enough variation for us to ignore requirement for Automatic Domain Randomisation in our work. However a more detailed work with ADR is required to prove generalisation of knowledge transfer between the pLLM agents. Finally, we assume perfect knowledge of object states, but the real world has messy perception, and our system needs to account for that.


\bibliographystyle{ACM-Reference-Format}
\bibliography{sample-base}


\end{document}